\documentclass{article}

\usepackage{hello}
\usepackage{fancyhdr} 
\fancypagestyle{firstpage}{
    \fancyhf{}
    \lhead{\includegraphics[width=4cm]{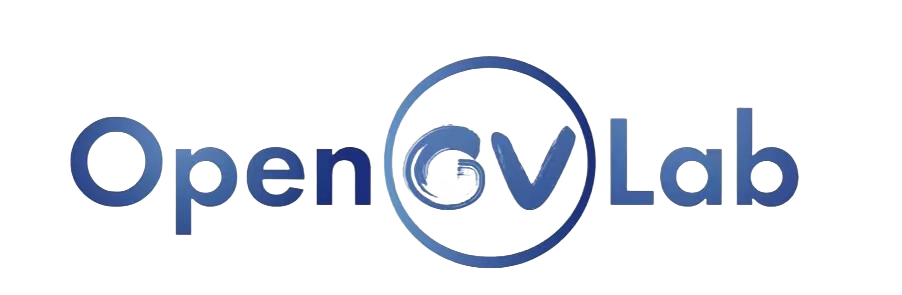}}
    \renewcommand{\headrulewidth}{0.4pt}
}

\usepackage[margin=1in]{geometry}
\usepackage[numbers]{natbib}

\usepackage[utf8]{inputenc} 
\usepackage[T1]{fontenc}    
\usepackage{url}            
\usepackage{booktabs}       
\usepackage{amsfonts}       
\usepackage{nicefrac}       
\usepackage{microtype}      
\usepackage{xcolor}         
\usepackage{multirow} 
\usepackage{array} 
\usepackage{makecell}
\usepackage{varwidth} 
\usepackage{endnotes}
\usepackage{microtype}      
\usepackage{enumitem}
\usepackage{amsmath}
\usepackage{amssymb}
\usepackage{multirow}
\usepackage{color}
\usepackage{xcolor}
\usepackage{colortbl}
\usepackage{subcaption}
\usepackage{pifont}
\usepackage{bbding}
\usepackage[most]{tcolorbox} 
\tcbuselibrary{listingsutf8}
\usepackage{CJK}

\definecolor{citecolor}{HTML}{0071BC}
\usepackage[backref=page,pagebackref=true,colorlinks,citecolor=citecolor,urlcolor=magenta,linkcolor=magenta]{hyperref}

\newcommand{\dummy}[1]{}

\makeatletter
\def\blfootnote{\xdef\@thefnmark{}\@footnotetext}
\makeatother

\title{	
Visual Embodied Brain: Let Multimodal Large Language Models See, Think, and Control in Spaces}

\author{
    \normalsize{}
    \textbf{Gen Luo$^{1*}$,~Ganlin Yang$^{3,1*}$,~Ziyang Gong$^{4,1*}$,~Guanzhou Chen$^{4,1*}$, 
    ~Haonan Duan$^6$, ~Erfei Cui$^{4,1}$}
    \\
    \normalsize{}\textbf{
    ~Ronglei Tong$^6$, ~Zhi Hou$^{1}$, ~Tianyi Zhang$^{7,1}$, ~Zhe Chen$^{8,1}$,  ~Shenglong Ye$^1$, ~Lewei Lu$^6$} \\
    \normalsize{}
   \textbf{~Jingbo Wang$^1$, ~Wenhai Wang$^1$, ~Jifeng Dai$^{2,1}$,
    ~Yu Qiao$^1$, ~Rongrong Ji$^5$, ~Xizhou Zhu$^{2\dag}$}  \\
     [1mm]
     \normalsize{}
    ~$^1$Shanghai AI Laboratory \quad $^2$Tsinghua University 
    \\ ~~\normalsize{}$^3$University of Science and Technology of China  ~~\normalsize{}$^4$Shanghai Jiao Tong University \\ ~~\normalsize{} $^5$Xiamen University    $^6$SenseTime Research  $^7$Zhejiang University $^8$Nanjing University\\
    [3mm]
    \normalsize{}
    Project Page: 
    \href{https://internvl.github.io/blog/2025-05-26-VeBrain/}{VeBrain}  
}
\date{}
 
\blfootnote{\noindent$^{*}$Equal contribution. $^{\dag}$Corresponding author.}

\begin{document}

\maketitle

\thispagestyle{firstpage} 
\begin{abstract}
The remarkable progress of Multimodal Large Language Models (MLLMs) has attracted increasing attention to extend them to physical entities like legged robot. This typically requires MLLMs to not only grasp multimodal understanding abilities, but also integrate visual-spatial reasoning and physical interaction capabilities. Nevertheless, existing methods struggle to unify these capabilities due to their fundamental differences. In this paper, we present the \textbf{V}isual \textbf{E}mbodied \textbf{Brain} (\textbf{VeBrain}), a unified framework for perception, reasoning, and control in real world. VeBrain reformulates robotic control into common text-based MLLM tasks in the 2D visual space, thus unifying the objectives and mapping spaces of different tasks. Then, a novel robotic adapter is proposed to convert textual control signals from MLLMs to motion policies of real robots.
From the data perspective, we further introduce VeBrain-600\textit{k}, a high-quality instruction dataset encompassing various capabilities of VeBrain.  In VeBrain-600\textit{k}, we take hundreds of hours to collect, curate and annotate the data, and adopt multimodal chain-of-thought (CoT) to mix the different capabilities into a single conversation.  Extensive experiments on 13 multimodal benchmarks and 5 spatial intelligence benchmarks demonstrate the superior performance of VeBrain to existing MLLMs like Qwen2.5-VL. When deployed to legged robots and robotic arms, VeBrain shows strong adaptability, flexibility, and compositional capabilities compared to existing methods. 
For example, compared to Qwen2.5-VL, VeBrain not only achieves substantial gains on MMVet by +5.6\%, but also excels in legged robot tasks  with +50\% average gains.  
\end{abstract}

\section{Introduction}
In recent years, Multimodal Large Language Models (MLLMs) have achieved significant progress in computer vision~\cite{VLM:GPT-4v, team2024gemini, bai2025qwen2, VLM:InternVL-1.5, luo2024mono}, continually pushing the boundaries of various vision-language tasks~\cite{Datasets:TextVQA,Datasets:ChartQA,Datasets:AI2D}. The next milestone of MLLMs is generally considered to be the migration of multimodal intelligence to physical entities, \emph{i.e.,} robotic arm~\cite{ahn2022can,gemini_robot,ji2025robobrain}, which could naturally possess the ability to perceive the surrounding world, reason in visual space, and actively engage with the environment. This requires  MLLMs to go beyond traditional multimodal understanding and incorporate both visual-spatial intelligence~\cite{yang2024thinking,azuma2022scanqa,ma2022sqa3d} and physical interaction capabilities~\cite{kim2024openvla,margolis2023walk}.

\begin{figure*}[t]
    \centering
    \includegraphics[width=1.\linewidth]{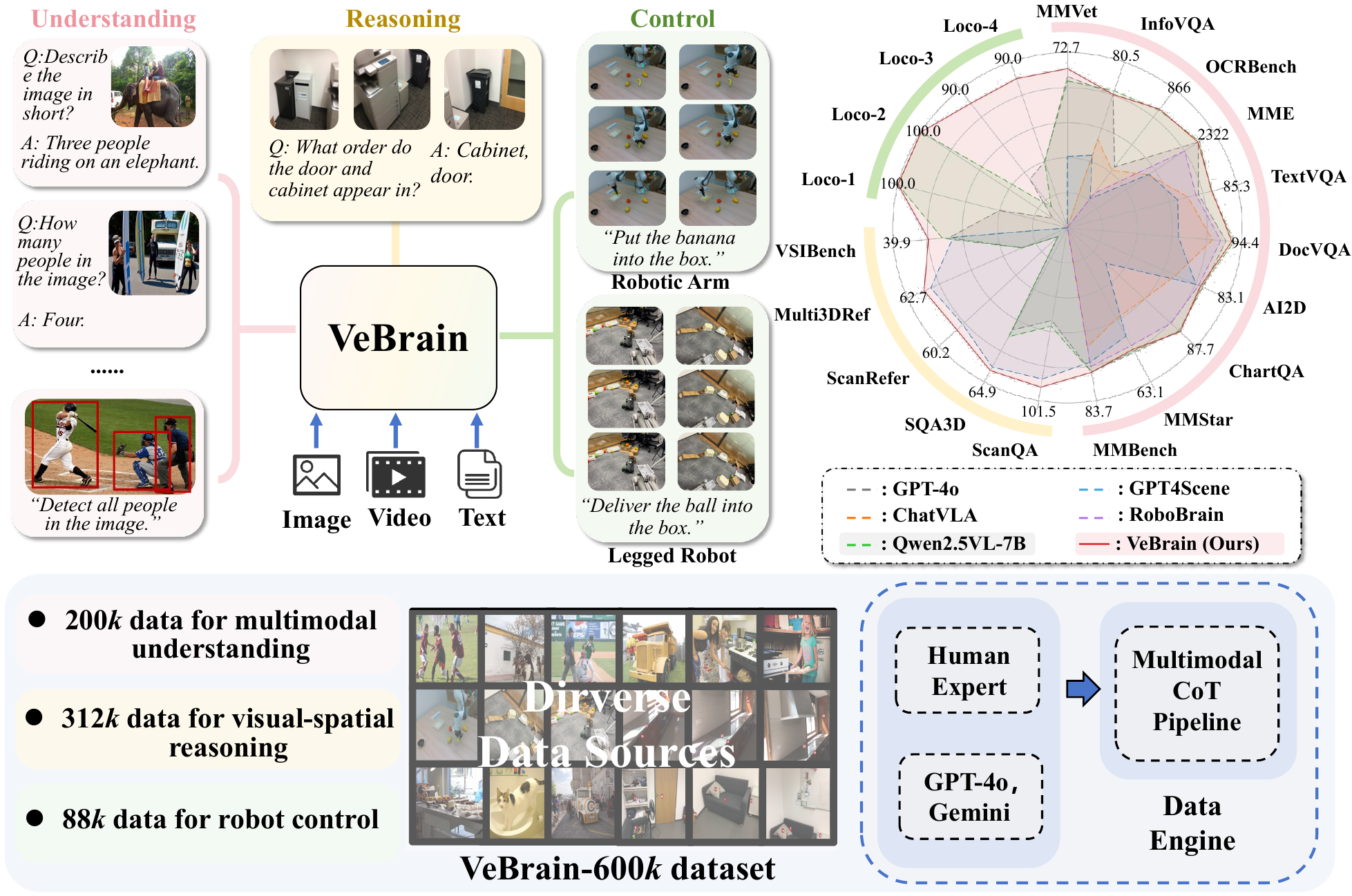}
    {
    \vspace{-1em}
    \caption{\textbf{Overview of VeBrain and VeBrain-600\textit{k}.} Compared to existing MLLMs, VeBrain achieves the best trade-off performance on benchmarks of multimodal understanding, visual-spatial reasoning, and robot control into one MLLM.  To support the unified training of VeBrain, VeBrain-600\textit{k} is built with a semi-automated data engine covering a variety of data sources and tasks.
    \label{fig:fig1}
    } 
    }
\end{figure*}

Nevertheless, existing MLLMs struggle to seamlessly incorporate these basic abilities into a single unified model. A notable approach is the vision-language-action (VLA) model~\cite{shridhar2022cliport, VLA:BC-Z, VLA:VIMA, VLA:RT-1,kim2024openvla,ding2024quar,VLA:chatvla,VLA:pi0}, where MLLMs are trained on large-scale robotic datasets to map multimodal observations into control policies. While effective for control tasks, the large-scale robotic learning of VLAs inevitably compromises their multimodal understanding capability~\cite{VLA:chatvla}. 
To compensate for this shortcoming, some attempts~\cite{VLA:vlmpc,mei2024quadrupedgpt} directly construct MLLM-based agents to control robots via text descriptions while preserving their multimodal reasoning abilities. However, due to the large task gap, their control accuracy and generalizability are still far from the requirements of real robots.

In this paper, we argue that the challenges of unifying these capabilities into MLLMs mainly arise from their inherent differences. Specifically, learning robot control demands a precise mapping from multimodal inputs to physical motion policies in the real world, \emph{e.g.,} vision-language-action (VLA) models~\cite{kim2024openvla,VLA:pi0}, which fundamentally differ from the cross-modal alignment objective of existing MLLMs in 2D visual space. This distinct objective makes it difficult for MLLMs to effectively balance these capabilities, leading to knowledge forgetting and task conflicts~\cite{VLA:chatvla}. Moreover, the community still lacks a suitable data recipe to seamlessly integrate and balance these capabilities in MLLMs, further exacerbating the problem.


To overcome these limitations, we propose the Visual Embodied Brain (VeBrain), a unified framework for perception, reasoning, and control in real world. The core idea of VeBrain is to formulate robot control as common text-based MLLM tasks in 2D visual space, thereby unifying the learning objectives across different capabilities for MLLMs. Specifically, robot control is decomposed into the tasks of keypoint detection~\cite{centernet,cornernet} and embodied skill recognition~\cite{mu2023embodiedgpt}. The former serves as a visual-spatial anchor encoding the movement signals of the robot, and the latter represents action commands for execution signals. Based on these control signals, we design a novel robotic adapter that converts them into motion policies in a dynamic and robust manner. With these designs, VeBrain can achieve efficient robot control while retaining the strong capabilities of MLLMs.

Based on VeBrain, we construct a high-quality instruction dataset to support the unified training, namely VeBrain-600\textit{k}. In addition to integrating open-source datasets, we construct an innovative semi-automated data engine that can generate task data requiring compositional capabilities. In particular, we first collect raw images and videos from public datasets and our collections, and then annotate the key information by human experts, \emph{e.g.,} keypoints.   Finally, capabilities like perception and reasoning are embedded into these data via multimodal chain-of-thought. Through these pipelines, VeBrain-600\textit{k} not only encourages MLLMs to jointly learn the basic capabilities of an embodied brain, but also maximizes its abilities in handling complex tasks.

To validate VeBrain, we conduct extensive experiments on over 20 benchmarks covering multimodal understanding, visual-spatial reasoning, and robot control. As shown in Fig.~\ref{fig:fig1}, experiments not only show the better performance of VeBrain than existing methods on multimodal tasks and spatial reasoning tasks, but also confirm its strong capabilities across diverse robotic tasks. For example,  VeBrain outperforms the state-of-the-art MLLM (Qwen2.5-VL~\cite{bai2025qwen2}) by +5.6\% on MMVet~\cite{yu2023mmvet} and achieves +0.2\% averaged gains of 13 multimodal benchmarks.  On legged robots and robotic arms, VeBrain also demonstrates superior adaptability than all baselines, \emph{e.g.,} +42.9\% average success rate against $\pi_0$.  In conclusion, our contributions are four folds:
\begin{itemize}
    \item We present VeBrain, a novel framework to unify multimodal understanding, visual-spatial reasoning, and robotic control. VeBrain formulates robot control as two MLLM tasks in 2D visual space, \emph{i.e.,} keypoint detection and embodied skill recognition, thereby avoiding potential conflicts with multimodal understanding.
    \item In VeBrain, we propose a novel robotic adapter that converts text-based control signals from MLLMs into the motion policies for real robots. The MLLM and the robotic adapter form a closed loop and jointly accomplish robotic tasks in a dynamic and robust manner.
    \item We present a high-quality instruction dataset covering the basic capabilities of VeBrain, namely VeBrain-600\textit{k}, which is constructed by human experts and semi-automated data engines. Besides, VeBrain-600\textit{k} includes a large amount of carefully constructed CoT data that can significantly benefit the compositional capabilities of VeBrain.
    \item  VeBrain is the first to outperform  existing MLLMs in terms of the average performance on 18 multimodal and spatial benchmarks. Moreover, its generalization ability in robotic control is also confirmed on 14 tasks of legged robots and robotic arms.
\end{itemize}

\section{Related Work}
\subsection{Multimodal Large Language Models}

The rapid advancement of large language models (LLMs)~\cite{openai2023gpt4, qwen, team2023gemini, touvron2023llama, TransF:LLaMA2, radford2018improving} has significantly propelled the development of multimodal large language models (MLLMs)~\cite{VLM:GPT-4v, team2024gemini, bai2025qwen2, VLM:InternVL-1.5, luo2024mono}. 
Models based on contrastive learning~\cite{clip, VLM:InternVL, jia2021scaling, fang2022eva} demonstrate strong open-world semantic alignment by matching image-text pairs, yet are inherently limited by their lack of generative capabilities.
By combining LLMs~\cite{brown2020language, radford2019language} with vision foundation models~\cite{TransF:MAE, clip, TransF:ViT}, MLLMs exhibit exceptional performance in various vision-language tasks~\cite{Datasets:MSCOCO, goyal2017vqav2, marino2019okvqa, Datasets:VizWiz, Datasets:TextVQA}. 
Concurrently, some works~\cite{VLP:Flamingo, awadalla2023openflamingo, li2024omnicorpus, lin2024vila} focus on cross-modal alignment through image-text interleaved training to boost contextual reasoning, while parallel efforts~\cite{chen2023shikra, lai2023lisa, 2023interngpt, peng2023kosmos2, wang2023allseeing} pioneer spatial-aware architectures for granular region understanding.
Despite many significant advances, these models mainly focus on perception and face challenges in real-world control scenarios, where embodied agents are expected to perceive and interact with physical environments~\cite{driess2023palme, mu2023embodiedgpt, cho2024spatially, liu2024coarse, zhu2024llava}. 
While building upon Qwen2.5-VL as our pretrained backbone, we uniquely employ a keypoint-based control mechanism, thus closing this critical operational gap prevalent in prior implementations.

\subsection{Vision-Language-Action Models}

Previous methods for robotic policy learning predominantly relied on reinforcement learning frameworks limited to narrow skill domains~\cite{levine2018learning, geng2023rlafford, zhu2017target, zhuang2023robot}. 
The impressive success of MLLMs, \emph{e.g.,} Flamingo~\cite{VLP:Flamingo}, BLIP-2~\cite{VLP:BLIPv2}, and LLaVA~\cite{VLM:LLaVA}, has inspired the emergence of vision-language-action models (VLAs). 
By fine-tuning on robotic interaction data, VLAs~\cite{shridhar2022cliport, VLA:BC-Z, VLA:VIMA, VLA:RT-1,VLA:interleave-VLA, VLA:dita} process visual observations and textual instructions through dedicated encoders, transforming these inputs into latent representations, which are subsequently decoded into executable action sequences.
Some approaches~\cite{VLA:R3M, VLA:MVP, VLA:VC-1, VLA:SMART,VLA:rovi} focus on the enhancement of individual components, such as pretrained visual representations, dynamics learning, and reasoning. 
Further studies like OpenVLA~\cite{kim2024openvla}, RT-2~\cite{brohan2023rt}, QUART~\cite{ding2024quar}, and NaVILA~\cite{cheng2024navila} demonstrate how VLAs enable precise action generation across diverse platforms, including robotic manipulators and quadruped robots. 
Recently, long-horizon tasks have gained increasing attention, giving rise to zero-shot agents~\cite{huang2022inner, liang2023code, driess2023palme} that serve as high-level planners responsible for task decomposition.
Nevertheless, existing VLAs struggle to match the understanding capabilities of MLLMs, which are significant for real-world applications.

\begin{figure*}[t]
    \centering
    \includegraphics[width=1.\linewidth]{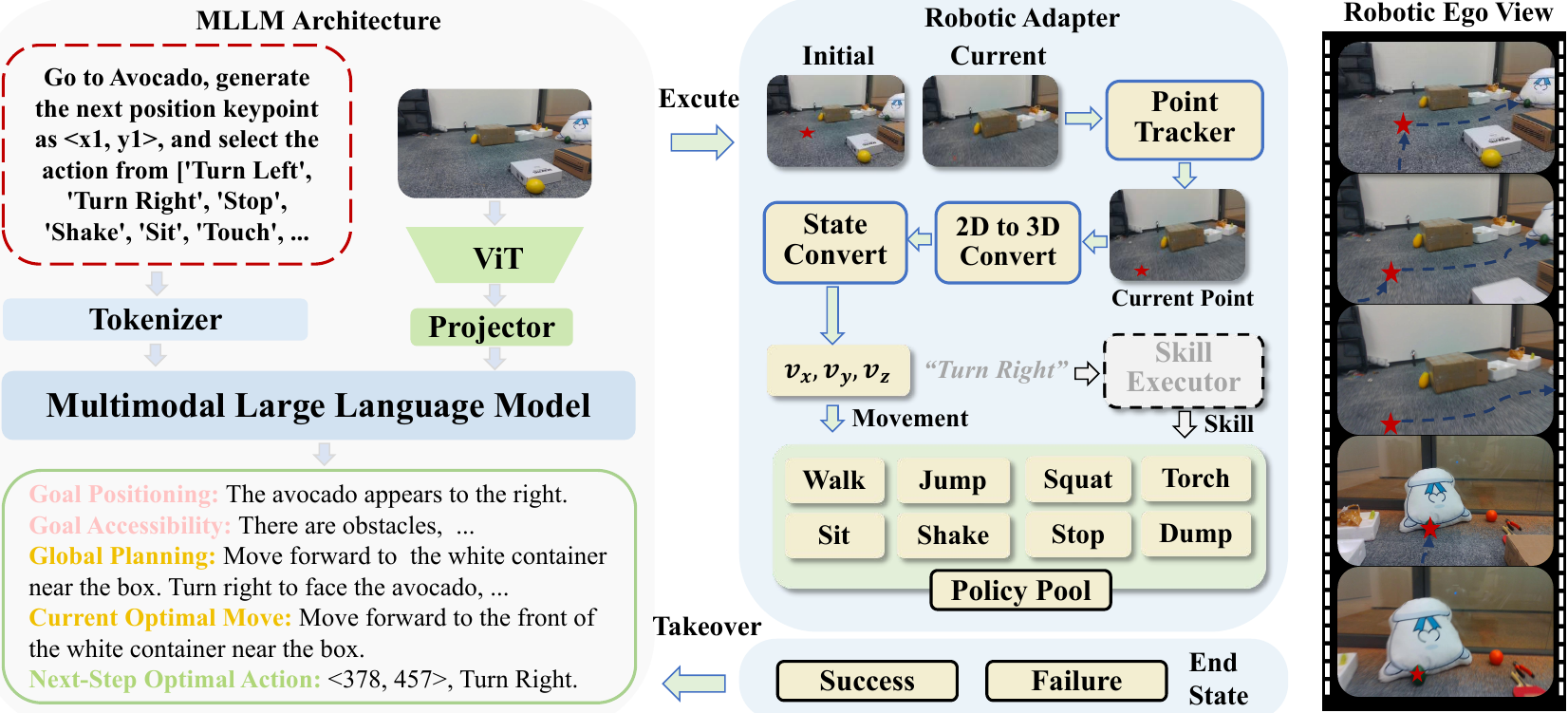}
    {
    \vspace{-1em}
    \caption{\textbf{Illustration of VeBrain architecture and robotic adapter.}  In VeBrain, the MLLM is capable of perception, thinking, and decision-making in common MLLM tasks. For robot control, an additional adapter is combined with the MLLM to achieve closed-loop control of the real robot. 
    \label{fig:fig2}
    }  
    }
\end{figure*}

\section{Method}

\subsection{Task Formulations} \label{sec:3.1} 
Previous endeavor~\cite{VLA:chatvla}   directly combines the formulation of vision-language-action (VLA) models and MLLMs, yet learning multimodal understanding and robotic control often negatively influence each other. To bridge this gap, we formulate these tasks based on three principles.  Firstly, for all tasks, we adopt the task modeling with the same input-output space, \emph{i.e.,} $p(t|x,y;\theta)$.  Here, $x \in \mathbb{R}^{T\times H \times W \times 3}$, $y\in \mathbb{R}^L$, $t \in \mathbb{R}^{N}$ and $\theta$ denote the visual input (images or video), the textual prompt, the answer and the model parameters, respectively. Secondly, robot control is defined as a common text-based MLLM task in 2D visual space, \emph{e.g.,}  point detection~\cite{bai2025qwen2}, thus reducing the learning difficulty. Thirdly, task-specific chain-of-thoughts are designed to help the model solve challenging problems step by step. Based on these principles, we describe the task definitions below.

\textbf{Multimodal Understanding \& Visual-Spatial Reasoning.} These tasks are already well-defined in existing MLLMs,  so we adopt a similar task formulation accordingly. Besides, we introduce the task-specific chain-of-thought (CoT) for these samples, where the MLLM is required to carefully analyze the problem and then give a response. Therefore, the  template of these tasks can be written as: 
`` prompt:  \texttt{<task description>}, answer:  \texttt{<thinking process> <answer>} '', where  <$\cdot$> denotes a slot  of specific textual content. Note that the thinking process is removed for easy samples.

\textbf{Robot Control.} The key challenge of robotic control lies in planning a given task step by step and then executing. As shown in Fig.~\ref{fig:fig2}, we design the environment perception, task planning, and current decision as a CoT process.  The current decision is decomposed into two sub-tasks for the MLLM, namely keypoint detection and embodied skill recognition.   Specifically, keypoints represent a set of end locations in an image that the robot should move to or interact with, while embodied skill denotes the action to be executed after the movement, \emph{e.g.,} clamp.  Note that the keypoints and embodied skills will be converted into executable programs via our robotic adapter.

\subsection{VeBrain Framework} 
As shown in Fig.~\ref{fig:fig2}, VeBrain consists of an MLLM for understanding, thinking and controlling, and a robotic adapter to convert the  MLLM decisions into executable policies.  These two parts are connected via a closed loop to enable dynamic and robust control.

\subsubsection{MLLM Architecture}  Our MLLM follows Qwen2.5-VL~\cite{bai2025qwen2}, which consists of the vision encoder, the projector, and the LLM. In particular, we use the optimized vision transformer (ViT) with a stride of 14 to extract visual features, where RMSNorm~\cite{zhang2019root}, SwiGLU~\cite{VLM:SwiGLU}, 2D-RoPE~\cite{bai2025qwen2}, and window attention~\cite{li2022exploring} are applied for improved efficiency and capability.  In particular,
given the input images $x\in\mathbb{R}^{t\times h\times w \times 3}$ and textual prompt $y\in\mathbb{R}^{l}$, the MLLM architecture of VeBrain can be defined by
\begin{equation}
    \begin{aligned}
        p= \mathcal{F_\text{llm}}(t_{N}| \mathcal{F_\text{v}}(x;\theta_v), \mathcal{F_\text{t}}(y), t_{0:N-1};\theta),
    \end{aligned} 
    \label{eq_arch}
\end{equation}
where  $p\in \mathbb{R}^{m}$ is the  next-token probability and $m$ denotes the vocabulary size. Here, $\mathcal{F_\text{v}}(\cdot)$ denotes the ViT and the MLP, and $\theta_v$ is their parameters. $\mathcal{F_\text{t}}(\cdot)$ is the textual tokenizer. $\mathcal{F_\text{llm}}(\cdot)$ and $\theta$ are the LLM and its parameters, respectively. $t_i$ denotes the \textit{i-th} predicted word. For tasks defined in Sec.~\ref{sec:3.1}, MLLM can directly predict the textual answer via Eq.~\ref{eq_arch}.

\subsubsection{Robotic Adapter} \label{Sec:robotic_controller}
As defined in Sec.~\ref{sec:3.1},   there is still a large gap between predictions of the MLLM and real-world deployment. Firstly, these 2D keypoints struggle to be directly applied to the real-world 3D scene. Secondly, the ego-view of legged robots changes in real time as they move, leading to the misalignment between keypoints and visual perspective. Thirdly, since the MLLM cannot perceive the robotic state, it is difficult to take control in time when unexpected situations occur. To overcome these limitations, we propose a robotic adapter that  consists of four modules, namely the point tracker, the movement controller, the skill executor, and the dynamic takeover.

\textbf{Point tracker.} In legged robots, the visual perspective from the ego-view camera differs when the robot moves. Thereby,  2D keypoints should also be updated accordingly to match the new perspective. To approach this target, we introduce a point tracking model, namely LocoTrack~\cite{LocoTrack}, where the keypoints of the current perspective are predicted in a real-time manner. 

\textbf{Movement controller.} The movement controller aims to generate the movement policy for a robot or robotic arm based on the 2D keypoints. Given the keypoints predicted by the MLLM, we can obtain the corresponding depth information from the RGBD camera. Then, these 2D keypoints are converted to 3D ones via a simple transformation of the calibration matrix. Based on these 3D points, we estimate the movement velocity of the robot and drive the underlying movement policy model.

\textbf{Skill executor.} Existing robots have accessed a variety of skills through pre-trained policies, \emph{e.g.,} walking~\cite{margolis2023walk} and jumping~\cite{li2023learning}, which are sufficient to allow the robot to interact with humans or spaces. Therefore, we collect these action policies and categorize them by name. Given the skill predicted by the MLLM, the skill executor will call the corresponding policy to accomplish the task.

\textbf{Dynamic takeover.} In the real world, the environmental uncertainty and the mistakes of the policy model often lead to the loss of targets. In this case, dynamic takeover aims to exchange control to the MLLM when the robotic adapter fails. In particular, the takeover happens when the key points disappear from view for several frames or when a subtask is done.



\subsection{VeBrain-600\textit{k} Data Engine} VeBrain-600\textit{k} contains extensive datasets covering basic capabilities of VeBrain. For each capability, our data collection strives to be as diverse as possible.
  
 \textbf{Data collection and annotation.}
For multimodal understanding, we construct a dataset with 200\textit{k} conversations, including data formats of 2D images, videos, and textual descriptions. Most of the data is collected from open-source datasets, such as ShareGPT4V~\cite{dataset:sharegpt4v}  and MMInstruct~\cite{dataset:mminstruct}, \emph{etc}. Another part of the data is generated by GPT4o and its CoT process is annotated by our pipeline.
 
For visual-spatial reasoning, we collect 312\textit{k} items from open-source datasets, \emph{i.e.,} GPT4Scene~\cite{qi2025gpt4scene}, and our annotated dataset. In our annotated dataset, given images from ScanNet~\cite{dataset:scannet}, we annotate images using two pipelines: 1) combining image frames and point-cloud snapshots to generate descriptive conversations via GPT-4o; 2) labeling information of counting, object size, and object distance via annotations from ScanNet and human experts.




 
For robot control, we collect a multimodal robot dataset with 88\textit{k} items from scratch, including locomotion and manipulation of legged robots and robotic arms. For data collection, 4 human experts take over 80 hours to collect video episodes and motion states of legged robots and robotic arms. Then, 5 human experts manually annotate the keypoints and actions of these episodes.

 

 

\textbf{Chain-of-thoughts generation.} Our CoT generation aims to embed different capabilities into one conversation via CoT. For multimodal understanding and visual-spatial reasoning, CoT content aims to integrate reasoning capabilities into these tasks. Thus, we design different CoT templates according to the task properties and generate CoT content using Gemini-2.0 and GPT-4o. For robot control, the CoT process further integrates contents of perception, reasoning, and control, which firstly describes the visual observation, then decomposes the task, and finally makes the control decision. More details about CoT generation are given in our appendix.

\begin{table*}[t]
    \centering
    \caption{\textbf{Ablation study of VeBrain on frameworks and data.} Our baseline follows VLM-PC~\cite{VLA:vlmpc} and replaces the MLLM with Qwen2.5-VL. }
           \vspace{-1em}
    \resizebox{\linewidth}{!}{
    \begin{tabular}{lcccccccc}
        \toprule
        \multirow{2}{*}{} & 
        \multicolumn{2}{c}{Multimodal Understanding} &
        \multicolumn{2}{c}{Visual-spatial Reasoning} & \multicolumn{2}{c}{Robot Control} &
        \multirow{2}{*}{Avg}\\
        
        \cmidrule(lr){2-3} \cmidrule(lr){4-5} \cmidrule(lr){6-7}
        & MMVet & MMBench & ScanQA (CIDEr) & VSI-Bench & Complex Find & Transporting & \\
        \midrule
        Qwen2.5-VL~\cite{bai2025qwen2}  & 67.1 & 83.5 & 62.7 & 35.9 & 0.0& 10.0& 43.2 \\
        + Robotic Adapter & 67.1 & 83.5 & 62.7 & 35.9 & 0.0 & 40.0 & 48.2\\
        + Control Data & 67.2 & 82.7 & 56.8 & 32.8 & 30.0 & 50.0 & 53.3\\
        + Spatial Reasoning Data & 64.7 & 82.1 & \textbf{102.2} & \textbf{40.3}  & 65.0 & 70.0 &  70.7 \\
        + Multimodal Understanding Data & \textbf{72.7} & \textbf{83.7} & 101.5 & 39.9 & \textbf{80.0} & \textbf{90.0} & \textbf{78.0} \\
        \bottomrule
    \end{tabular}}
    \label{tab:ablation_1} 
\end{table*}

\begin{table}[t]
    \centering
    \caption{\textbf{Comparison of VeBrain with two common frameworks.} For VLA, the training data remains the same as VeBrain, except that the robot annotations are replaced with motion policies~\cite{ding2024quar}. }
           \vspace{-1em}
        \resizebox{\linewidth}{!}{
    \begin{tabular}{llcccccccc}
        \toprule
        \multirow{2}{*}{Forms} & \multirow{2}{*}{\makecell{Control\\Signal}} & \multirow{2}{*}{\makecell{Robotic\\Adapter}} &
        \multicolumn{2}{c}{Multimodal Understanding} & \multicolumn{2}{c}{Visual-spatial Reasoning} & \multicolumn{2}{c}{Robot Control} &
        \multirow{2}{*}{Avg} \\
        \cmidrule(lr){4-5} \cmidrule(lr){6-7} \cmidrule(lr){8-9}
        & && MMVet & MMBench & ScanQA (CIDEr) & VSI-Bench & Complex Find & Transporting &\\
        \midrule
        MLLM~\cite{VLA:vlmpc}  & Text   &     \XSolidBrush    & 67.1 & 83.5 & 62.7 & 35.9 & 10.0 & 20.0 & 46.5\\
        VLA~\cite{ding2024quar}   & Action policy & \XSolidBrush   & 50.8  &  73.4   &  55.1      &  29.8  & 50.0 & 30.0  & 48.2\\
        VeBrain (Ours)  & Points \& Action       & \Checkmark    &   \textbf{72.7}   &   \textbf{83.7}   &  \textbf{101.5}      &   \textbf{39.9}     & \textbf{80.0} & \textbf{90.0} & \textbf{78.0}\\
        \bottomrule
    \end{tabular} } 
    \label{tab:ablation_2} 
\end{table}

\begin{table*}[t!]
\centering
\setlength\tabcolsep{1pt}
\renewcommand{\arraystretch}{1.15}
\caption{\textbf{Comparison with existing MLLMs and VLAs on general MLLM benchmarks.} For MME, we sum the perception and cognition scores.   Avg denotes the  normalized average performance of MLLM benchmarks and VQA benchmarks. The highest score among open-source MLLMs and VLAs are colored in bold.}
\vspace{-1em}
\resizebox{\linewidth}{!}{
\begin{tabular}{l|ccccccccccccc|c}
\toprule
\textbf{Model} & \textbf{MMBench}& \textbf{MMVet} & \textbf{MMMU} & \textbf{MME} & \textbf{MMStar} & \textbf{RWQA} &   \textbf{TextVQA}  & \textbf{DocVQA} & \textbf{AI2D} & \textbf{ChartQA} & \textbf{InfoVQA} & \textbf{SEED2+} & \textbf{OCRBench} & \textbf{Avg} \\
\midrule
\rowcolor{gray!15} \multicolumn{15}{l}{$\blacktriangledown$ \emph{Closed-source MLLMs:}}  \\
GPT-4o~\cite{gpt4o}                             & 83.4             & 69.1           & 69.1          & 2328         & 64.7            & 75.4                 & 77.4             & 91.1            & 84.6          & 86.7             & 80.7             & 72.0                 & 736               &     76.5 
         \\
Claude-3.5 Sonnet \cite{claude3.5}                  & 82.6             & 70.1           & 68.3          & 1920         & 65.1            & 60.1                 & 76.5             & 95.2            & 81.2          & 90.8             & 74.3             & 71.7                 & 788               &     74.6 
         \\
Gemini-1.5-Pro~\cite{team2024gemini}                     & 73.9             & 64.0           & 62.2          & $-$          & 59.1            & 67.5                 & 78.8             & 93.1            & 88.4          & 87.2             & 81.0             & 70.8                  & 754               &      $-$        \\

\midrule
\rowcolor{gray!15} \multicolumn{15}{l}{$\blacktriangledown$ \emph{Open-source MLLMs:}}  \\
MiniCPM-V2.6-8B \cite{yao2024minicpm}                    & 81.5             & 60.0           & 49.8          & \textbf{2348}          & 57.5            & 65.0                 & 80.1             & 90.8            & 82.1          & 82.4             & $-$              & 65.7                  & 852               &     $-$         \\
LLaVA-One-Vision-7B~\cite{li2024llava}                & 80.9             & 57.5           & 48.8          & 1998          & 61.7            & 66.3                 & $-$              & 87.5            & 81.4          & 80.0             & 68.8             & $-$                  & 622               &      $-$        \\
Eagle2.5-8B~\cite{chen2025eagle}                        & $-$             & 62.9           & 55.8          & $-$          & \textbf{66.2}            & \textbf{76.7}                 & 83.7             & 94.1            & \textbf{84.5}          & 87.5             & 80.4             & $-$                  & \textbf{869}               &    $-$          \\
InternVL2.5-8B~\cite{chen2024expanding}                     & \textbf{84.6}             & 62.8           & 56.2          & 2344         & 62.8            & 70.1                 & 79.1             & 93.0            & \textbf{84.5}          & 84.8             & 77.6             & 69.7                 & 822               &      75.0 
        \\
Qwen2.5-VL-7B~\cite{bai2025qwen2}                       & 83.5             & 67.1           & \textbf{58.6}          & 2347         & 63.9            & 68.5                 & 84.9             & \textbf{95.7}            & 83.9          & 87.3             & \textbf{82.6}             & 70.4                 & 864               &     76.9 
         \\
\midrule
\rowcolor{gray!15} \multicolumn{15}{l}{$\blacktriangledown$ \emph{Vision-language-action Models:}}   \\

OpenVLA~\cite{kim2024openvla}                            & 0                & 0              & 0             & 0            & 0               & 0                    & 0                & 0               & 0             & 0                & 0                & 0                    & 0                 &    0          \\
ECoT~\cite{zawalski2024robotic}                               & 0                & 0              & 5.4           & 0            & 0               & 0                    & 0                & 0               & 0             & 0                & 0                & 0                    & 12                &      1.3        \\
DiVLA~\cite{wen2024diffusion}                              & $-$              & $-$            & 17.2          & 187        & 21.1            & 25.2                 & 7.5              & 15.2            & 43.1          & 17.2             & 14.7             & $-$                  & 294               &       $-$       \\
ChatVLA~\cite{VLA:chatvla}                            & 69.0             & $-$            & 37.4          & 1435       & 47.2            & 57.0                 & 71.2             & 83.3            & 67.6          & 59.9             & 53.3             & $-$                  & 729               &        $-$      \\
RoboBrain~\cite{ji2025robobrain}                          & 80.4            & $-$            & 49.0          & 2084         & 61.2            & 68.8                & 75.9             & 88.0            & 82.0          & 80.5             & $-$              & $-$                  & 677               &     $-$         \\
\midrule
VeBrain  (Ours)                            & 83.7             & \textbf{72.7}           &      56.6         & 2322         & 63.1            & 71.0                 & \textbf{85.3}             & 94.4            & 83.1          & \textbf{87.7}             & 80.5             & \textbf{70.7}                 & 866               &    \textbf{77.1}          \\
\bottomrule
\end{tabular}
}

\label{tab:mllm_benchmark} 
\end{table*}

\begin{table}[t]
    \centering
     \caption{\textbf{Comparison of VeBrain and existing MLLMs on four 3D spatial benchmarks.}}
    \setlength{\tabcolsep}{2pt}
    \vspace{-1em}
    \resizebox{\linewidth}{!}{
    \begin{tabular}{lccccccccccc}
        \toprule
        \multirow{2}{*}{Models} & \multicolumn{5}{c}{ScanQA (val)} & \multicolumn{2}{c}{SQA3D (val)} &
        \multicolumn{2}{c}{ScanRefer} &
        \multicolumn{2}{c}{Multi3DRef} \\
        \cmidrule(lr){2-6} \cmidrule(lr){7-8} \cmidrule(lr){9-10} \cmidrule(lr){11-12}
        & BLEU-1 & BLEU-4 & METEOR & ROUGE & CIDEr & EM-1 & EM-R1 & Acc@0.25 & Acc@0.50 & all F1@0.25 & all F1@0.50\\
        \midrule
        \rowcolor{gray!15} \multicolumn{12}{l}{$\blacktriangledown$ \emph{3D MLLMs:}}   \\
        Chat-3D v2~\cite{huang2023chat3d}                  & 38.4   & 7.3    & 16.1   & 40.1  & 77.1  & $-$            & $-$            & $-$           & $-$           & $-$            & $-$           \\
3D-LLM~\cite{hong20233d}                      & 39.3   & 12.0   & 14.5   & 35.7  & 69.4  & $-$            & $-$            & $-$           & $-$           & $-$            & $-$           \\
LEO~\cite{huang2023embodied}                         & $-$    & 11.5   & 16.2   & 39.3  & 80.0  & 50.0           & 52.4           & $-$           & $-$           & $-$            & $-$           \\
3DVG-Transformer~\cite{zhao20213dvg}            & $-$    & $-$    & $-$    & $-$   & $-$   & $-$            & $-$            & 47.6          & 34.7          & $-$            & 25.5          \\
M3DRef-CLIP~\cite{zhang2023multi3drefer}                 & $-$    & $-$    & $-$    & $-$   & $-$   & $-$            & $-$            & 51.9          & 44.7          & 42.8           & 38.4          \\
Chat-Scene~\cite{huang2023chat}                  & 43.2   & 14.3   & 18.0   & 41.6  & 87.7  & 54.6           & 57.5           & 55.5          & 50.2          & 57.1           & 52.4          \\\midrule
\rowcolor{gray!15} \multicolumn{12}{l}{$\blacktriangledown$ \emph{2D MLLMs:}}   \\
GPT-4o~\cite{gpt4o}               & 27.3   & 7.3    & 12.5   & 37.7  & 59.1  & 42.7           & 46.4           & 5.4           & 5.1           & 21.1           & 19.9          \\
Qwen2.5-VL-7B~\cite{bai2025qwen2}               & 25.7   & 9.8    & 12.3   & 33.1  & 62.7  & 45.8           & 49.5           & 5.4           & 5.1           & 21.1           & 19.9          \\
GPT4Scene-HDM~\cite{qi2025gpt4scene} & 44.4   & 15.5   & 18.9   & 46.5  & 96.3  & 60.6           & 63.3           & 62.6          & 57.0          & 64.5           & 59.8         \\ \midrule
VeBrain (Ours) & \textbf{47.7} & \textbf{17.3} & \textbf{20.1} & \textbf{48.2}	& \textbf{101.5} & \textbf{61.6} & \textbf{64.9} & \textbf{66.4}	& \textbf{60.2}	& \textbf{67.8} & \textbf{62.7} \\
        \bottomrule
    \end{tabular}}
    \label{tab:3d_benchmark}
\end{table}

\begin{table}[t]
    \centering
    \caption{\textbf{Comparison of VeBrain and existing MLLMs on VSI benchmark.}}
    \setlength{\tabcolsep}{2pt}
    \vspace{-1em}
    \resizebox{\linewidth}{!}{
    \begin{tabular}{lccccccccc}
        \toprule
        Models & AVG & Obj. Count & Abs. Dist & Obj. Size & Room. Size & Rel. Dist & Rel. Dir & Router Plan & Appr. Order\\
        \midrule
        GPT-4o~\cite{gpt4o}               & 34.0                     & 46.2       & 5.3       & 43.8      & 38.2       & 37.0      & 41.3     & 31.5        & 28.5        \\
Gemini-1.5 Pro~\cite{team2024gemini}       & 45.4                     & 56.2       & 30.9      & 64.1      & 43.6       & 51.3      & 46.3     & 36.0        & 34.6        \\
\midrule
VILA-1.5-8B~\cite{lin2024vila}          & 28.9                     & 17.4       & 21.8      & 50.3      & 18.8       & 32.1      & 34.8     & 31.0        & 24.8        \\
LongVA-7B~\cite{zhang2024long}            & 29.2                     & 38.0       & 16.6      & 38.9      & 22.2       & 33.1      & \textbf{43.3}     & 25.4        & 15.7        \\
LLaVA-NeXT-Video-7B~\cite{zhang2024llavanextvideo}  & 35.6                     & 48.5       & 14.0      & 47.8      & 24.2       & \textbf{43.5}      & 42.4     & \textbf{34.0}        & 30.6        \\
LLaVA-OneVision-7B~\cite{li2024llava}   & 32.4                     & 47.7       & 20.2      & 47.4      & 12.3       & 42.5      & 35.2     & 29.4        & 24.4        \\
Qwen2.5-VL-7B~\cite{bai2025qwen2}         & 35.9 & 40.6       & 20.3      & 50.5      & 35.7       & 37.0      & 40.5     & 29.9        & \textbf{32.4}      \\  \midrule
VeBrain (Ours) & \textbf{39.9} & \textbf{59.5}	& \textbf{31.2} &	\textbf{53.7} &	\textbf{40.3} &	41.0 &	36.9 & 29.9	& 26.2 \\
        \bottomrule
    \end{tabular}}
    \vspace{0.5em}
    \label{tab:vsi_benchmark}
    \vspace{-1em}
\end{table}

\begin{table}[t]
\centering
\caption{\textbf{ Performance comparison on 7 tasks of legged robot.} We report the success rate over ten trials for each task. Task details are given in the appendix.}
\vspace{-1em}
\resizebox{\linewidth}{!}{
\begin{tabular}{lccccccccccc}

\toprule
\multirow{3}{*}{Models} & \multirow{3}{*} {\makecell{Robotic\\Adapter}} & \multicolumn{3}{c}{Easy} & \multicolumn{3}{c}{Middle} & \multicolumn{1}{c}{Hard} & \multirow{3}{*}{Overall}\\
\cmidrule(lr){3-5} \cmidrule(lr){6-8} \cmidrule(lr){9-9}

& & Find & Track & Interaction & \makecell{Complex\\Find} & \makecell{Complex\\Interaction}& Transport & \makecell{Complex\\Transport} &    \\
\midrule
VLA~\cite{ding2024quar} &   \XSolidBrush  & 70.0 & 30.0 & 20.0 & 50.0& 15.0& 30.0&10.0 & 32.1\\

VLM-PC~\cite{VLA:vlmpc} &   \XSolidBrush  & 60.0 & 40.0 & 10.0 & 10.0 & 5.0 &20.0 & 0.0 & 20.7\\

GPT-4o~\cite{gpt4o} &   \Checkmark & 40.0 & 10.0 & 35.0 & 10.0 & 5.0 & 40.0 & 10.0 & 21.4  \\

Gemini-1.5 Pro~\cite{team2024gemini} & \Checkmark & 20.0 & 0.0 & 20.0 & 0.0 & 0.0 & 30.0 & 0.0 & 10.0  \\

Qwen2.5-VL-7B~\cite{bai2025qwen2} & \Checkmark & \textbf{100.0 }& \textbf{100.0} & 15.0 & 20.0 & 10.0 & 40.0 & 10.0 & 42.1\\ \midrule

VeBrain (Ours) & \Checkmark &\textbf{ 100.0 }& \textbf{100.0} & \textbf{90.0} & \textbf{80.0} & \textbf{85.0} &\textbf{ 90.0 }& \textbf{60.0} & \textbf{86.4} \\
\bottomrule
\end{tabular}}
\vspace{0.5em}
\label{tab:legged_benchmark}
    \vspace{-1em}
\end{table}

\begin{table}[t]
\centering
\caption{\textbf{Performance comparison on 7 tasks of robotic arm.} All models are allowed to see 10 demonstrations per task during training. Task details are given in the appendix.}
\setlength{\tabcolsep}{3.5pt}
\begin{tabular}{lccccccccccccc}
\toprule
\multirow{2.5}{*}{Model} &\multirow{2.5}{*}   &\multicolumn{2}{c}{Move into Box} && \multicolumn{2}{c}{Move out of Box} & & \multirow{2.5}{*} {\makecell{Open\\Drawer}}& &\multicolumn{2}{c}{Long-Horizon} &&\multirow{2.5}{*}{Overall}\\
\cmidrule{3-4}\cmidrule{6-7}\cmidrule{11-12}
                       & & Banana                        & Pepper                  &      & Carrot                        & Kiwifruit                   & & &  & Carrot            & Pepper            \\
\midrule
OpenVLA~\cite{kim2024openvla}    &   &  0.0   &   10.0   &       &   10.0  &   20.0  & &40.0 & & 0.0 & 0.0 &&11.4     \\ 
$\pi_0$~\cite{VLA:pi0}    &   &0.0  &  30.0  & &  \textbf{90.0 }&  50.0  & & 50.0 &   & 0.0 & 0.0 &&31.4      \\ 
VeBrain (Ours)      &      &  \textbf{70.0} & \textbf{70.0} &   & \textbf{90.0 }&\textbf{ 60.0} & & \textbf{90.0} && \textbf{60.0 }& \textbf{ 80.0}&&\textbf{74.3}   \\       
\bottomrule
\end{tabular}
\vspace{0.5em}
\label{tab:arm_benchmark} 
\end{table}

\textbf{Quality verification.}  
During the robot data collection process, three experts carefully reviewed each video to ensure that the objects were within the robot's field of view. 
For CoT generation, we adopt a cross-model validation pipeline. In particular,  we employ Gemini-2.0 as the reference model to assess the logical and physical plausibility of CoT data generated by GPT-4o.  
Finally, we randomly select 10\% of the data for manual inspection by 5 human experts, and only 5.3\% of them are further excluded, demonstrating the reliability of our data generation pipeline. 

\section{Experiments}
\subsection{Evaluation Datasets}
For multimodal capability, we evaluate VeBrain
and existing MLLMs on 13 comprehensive multimodal benchmarks. 
Specifically,  MLLM benchmarks encompass MMBench-EN \textit{test}~\cite{Datasets:MMBench}, MMVet~\cite{Datasets:MM-vet}, MMMU \textit{val}~\cite{Datasets:MMMU}, MME~\cite{Datasets:MME}, 
MMStar~\cite{chen2024we} and
RealWorldQA~\cite{VLM:Grok-1.5}.
Visual question answering benchmarks include TextVQA \textit{val}~\cite{Datasets:TextVQA},  DocVQA \textit{test}~\cite{Datasets:DocVQA}, AI2D \textit{test}~\cite{Datasets:AI2D}, ChartQA \textit{test}~\cite{Datasets:ChartQA}, InfoVQA \textit{test}~\cite{mathew2022infographicvqa}, 
SEED-Bench-2-Plus~\cite{li2024seed}
and OCRBench~\cite{liu2023ocrbench}.
 For visual-spatial reasoning, we evaluate VeBrain on five benchmarks, including ScanQA \textit{val}~\cite{azuma2022scanqa}, SQA3D \textit{val}~\cite{ma2022sqa3d},  ScanRefer~\cite{chen2020scanrefer}, Multi3DRef~\cite{zhang2023multi3drefer}, and VSI-Bench~\cite{yang2024thinking}. 
 For robot control, evaluation benchmarks are built from our self-collected scenes, as described in the appendix.

\subsection{Implementation Details}
VeBrain is implemented based on Qwen2.5-VL-7B~\cite{bai2025qwen2}, which is fine-tuned on our VeBrain-600\textit{k} dataset with a learning rate of 5e-6 for one epoch, taking approximately 2 days on 32 NVIDIA A100 GPUs. After training, VeBrain is deployed to a Tesla A100 GPU in the cloud, running at 0.5Hz. The tracking model is deployed on the NVIDIA Jetson Orin with a 15Hz running frequency. Our legged robot (Unitree Go2) is equipped with an ego-view RGBD camera (RealSense D435i) and a Jetson AGX Orin platform. The robotic arm consists of a 7-DoF tabletop Franka Emika Panda robot arm and a Robotiq 2F-85 gripper, which is equipped with an RGBD camera in a third-person view. VeBrain and these robots communicate via a wireless network. More details are given in the appendix.

\subsection{Quantitative Results}

\textbf{Ablation studies.}  In Tab.~\ref{tab:ablation_1}, we validate the effectiveness of the data and architecture of VeBrain.  From this table, the first observation is that despite the promising performance on multimodal understanding, existing MLLMs often fall short in visual-spatial reasoning and robot control, \emph{i.e.,} 0\% success rate on the ``Complex Find'' task. After equipping the model with our robotic adapter, the success rate of Qwen2.5-VL on two robot control tasks is obviously improved.  Another observation is that after fine-tuning on control data, the multimodal capabilities of VeBrain are well preserved, greatly confirming the design of VeBrain. In addition, each type of data makes a significant contribution to the corresponding capabilities, \emph{e.g.,} +7.5\% of spatial reasoning data on VSI-Bench~\cite{yang2024thinking}.   

In Tab.~\ref{tab:ablation_2}, we compare VeBrain with two common frameworks, \emph{i.e.,} MLLM~\cite{VLA:vlmpc} and VLA~\cite{ding2024quar}. From these results, we find that  MLLM struggles to directly control the robot on two tasks due to its weak control capabilities. In contrast, VLA can perform well on robot control tasks, but it greatly sacrifices multimodal abilities, \emph{e.g.,} -16.3\% on MMVet compared to MLLM. Compared to these frameworks, VeBrain achieves the best trade-off performance on all tasks, yielding up to +31.5\% average gains against other frameworks. These results not only validate the shortcomings of existing frameworks in unifying multimodal understanding, visual-spatial reasoning, and robot control, but also confirm the effectiveness of each design in VeBrain. 

\begin{figure*}[t]
    \centering
    \includegraphics[width=1.\linewidth]{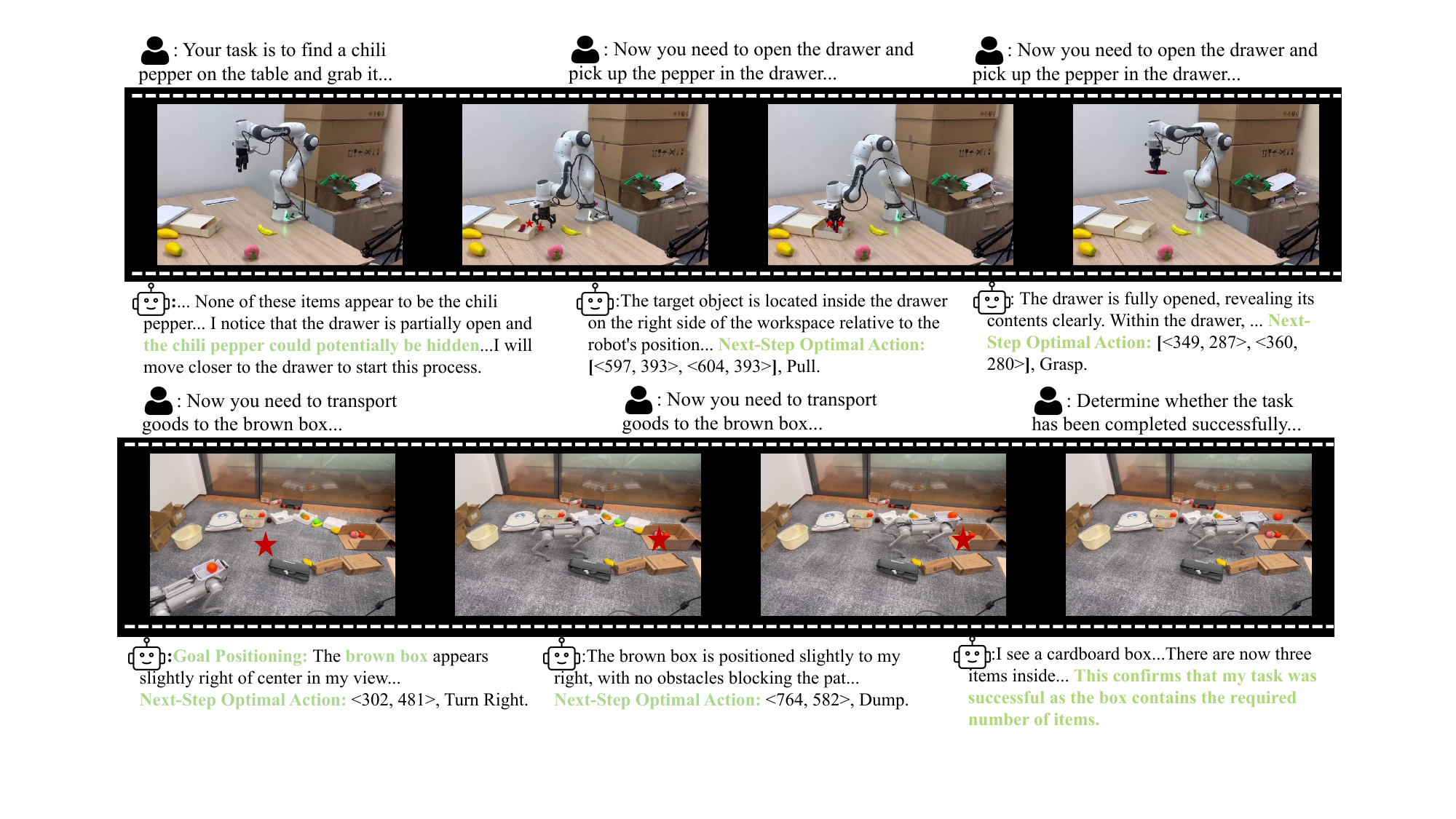}
    {
    \vspace{-1.5em}
    \caption{\textbf{Visualization of VeBrain on robotic arm and legged robot.} VeBrain demonstrates compositional capabilities in handling complex robotic tasks. Due to space limitations, most of the text regarding the thinking process is omitted in the figure.
    \label{fig:vis}
    }  
    }
\end{figure*}

\textbf{Results of multimodal understanding.} In Tab.~\ref{tab:mllm_benchmark}, we observe that existing VLA models perform inferiorly on these multimodal benchmarks, while OpenVLA completely loses the multimodal capabilities. Among them, RoboBrain~\cite{ji2025robobrain} integrates a large amount of multimodal understanding data in its training and achieves better results than other VLA models.  However, compared to advanced MLLMs like Qwen2.5VL, RoboBrain obviously falls short in OCR and chart benchmarks, \emph{e.g.,} -6.8\%  on ChartQA and -187 on OCRBench. In contrast, VeBrain demonstrates the comprehensive advantages across 13 MLLM benchmarks, \emph{e.g.,} +6.4\% against RoboBrain on DocVQA. Besides, VeBrain performs better than advanced open-source and closed-source MLLMs on most benchmarks, 
as proved by +5.6\% on MMVet and the best normalized average performance of 77.1,
suggesting its superior multimodal capabilities.   Considering the much smaller size of VeBrain than closed-source MLLMs, these results further confirm the strong multimodal abilities of VeBrain.

\textbf{Results of visual-spatial reasoning.}  Tab.~\ref{tab:3d_benchmark} first demonstrates the results of four 3D spatial benchmarks. In this table, the models require strong 3D spatial perception and reasoning capabilities to answer different types of questions. Therefore, most previous approaches~\cite{huang2023chat,zhang2023multi3drefer} adopt the 3D-based MLLM structure and achieve promising results on the four benchmarks.  In comparison, directly transferring 2D MLLMs to these tasks leads to poor performance, \emph{e.g.,}  -50.1 Acc@0.25 of Qwen2.5-VL-7B~\cite{bai2025qwen2}, suggesting their shortcoming in 3D spatial understanding and reasoning.  Compared to these methods,  GPT4Scene-HDM~\cite{qi2025gpt4scene} achieves better results via the video-based 2D MLLM and object markers.   However, as a specialist model, GPT4Scene-HDM struggles to be applied to common 2D multimodal tasks.  In contrast, VeBrain, as a generalist MLLM, can even outperform GPT4Scene-HDM on all tasks, \emph{e.g.,} +5.2 CIDEr on ScanQA \textit{val}, greatly validating its generalizability.  Tab.~\ref{tab:vsi_benchmark}  further diagnoses the visual-spatial reasoning abilities of existing MLLMs and VeBrain. From it, we can see that VeBrain outperforms all existing MLLMs on the VSI benchmark in terms of the average score, \emph{e.g.,} +4.0\% against the Qwen2.5-VL-7B. Compared to much larger MLLMs like GPT-4o, VeBrain can also perform even better. 
Considering the great challenge of the VSI benchmark, these results further confirm the spatial reasoning capabilities of VeBrain.  

\textbf{Results of robot control.} 
In Tab~\ref{tab:legged_benchmark}, we compare VLA, MLLMs, and VeBrain on seven tasks of the legged robot.  From it, we find that existing VLA~\cite{ding2024quar} and MLLMs~\cite{gpt4o, team2024gemini, bai2025qwen2} have difficulty in directly accomplishing most tasks like ``Interaction'' and ``Transport''.  Among them, Qwen2.5-VL  with our robotic adapter achieves the best results. However, when adapting it to harder tasks like ``Complex Find'', their success rate drops significantly, \emph{e.g.,} 20\% success rate.  These tasks typically require compositional capabilities like spatial reasoning and embodied control, which common MLLMs are not good at. Compared to these methods, VeBrain unifies these capabilities and achieves significantly better results on various complex tasks of legged robots. For example, VeBrain outperforms all models by 50\% on the long-horizon task ``Complex Transport''.  Similar merits of VeBrain can also be witnessed in the robotic arm. As shown in Tab.~\ref{tab:arm_benchmark}, common VLAs~\cite{kim2024openvla,VLA:pi0} demonstrate limited success rate in most  manipulation tasks, \emph{e.g.,} 30\% success rate of $\pi_0$~\cite{VLA:pi0} on ``Move Pepper into Box''. In long-horizon tasks, the success rate of $\pi_0$ further drops to 0\%.  Compared with these methods, VeBrain achieves the highest success rate in all tasks. In the most challenging task,  VeBrain outperforms $\pi_0$ by up to 80\%,  which further validates its effectiveness in robot control.

\subsection{Visualizations} In Fig.~\ref{fig:vis}, we visualize the results of VeBrain on real robots, \emph{i.e.,} robotic arm and legged robot.  From this figure, we  see that VeBrain can handle complex robotic tasks through compositional capabilities. For example, when asked to find a hidden chili pepper, VeBrain can correctly guess its potential location and then step through the steps to catch it. This requires not only control abilities, but also excellent perception and reasoning capabilities.  Similar merits are also reflected in the second example, where VeBrain further determines whether the goods have been delivered to the destination.


\section{Conclusion}
In this paper, we propose VeBrain, a visual embodied brain that unifies multimodal understanding, visual-spatial reasoning, and robot control. In VeBrain, robot control is formulated as multimodal tasks of key point detection and embodied skill recognition, and a novel robotic adapter is proposed to convert these signals of VeBrain to motion policies.  Based on this framework, we propose the VeBrain-600\textit{k} dataset to enable the learning of different capabilities. Extensive experiments on over 20 benchmarks demonstrate the superior performance of VeBrain than existing methods in tasks of three capabilities. In particular, VeBrain is the first to outperform the state-of-the-art MLLM in most multimodal tasks while retaining strong capabilities of spatial reasoning and robot control. 
\paragraph{Acknowledgments.}
This work was supported by the National Key R\&D Program of China (NO. 2022ZD0161302), the National Natural Science Foundation of China (No. 623B2088) and the China Postdoctoral Science Foundation (No. 2024M761548).

\bibliographystyle{ieeenat_fullname}
{
    \small
    \bibliography{neurips25}
}

\newpage

 \appendix
\section{More Implementation Details}
\subsection{Training Details} 

Our VeBrain is optimized in a fully supervised finetuning (SFT) manner based on Qwen2.5-VL-7B-Instruct~\cite{bai2025qwen2}. We let the whole parameters in large language model (Qwen2.5 LLM Decoder) trainable while keeping others frozen.
The detailed training configuration is located in Tab.\ref{tab:training_hype}.

\begin{table}[ht]
    \centering
     \caption{\textbf{Hyper-parameters used in the training of VeBrain.}}
    \setlength{\tabcolsep}{30pt}
    \footnotesize
    \begin{tabular}{lc}
        \toprule
        Configurations & Values  \\
        \midrule
        LLM sequence length & 16384 \\
        Max pixel length & 12845056 \\
        Freeze vision tower & True \\
        Freeze multimodal projector & True\\
        Freeze language model & False\\
        Optimizer & AdamW \\
        Optimizer hyperparameters & $\beta_1=0.9$, $\beta_2=0.999$, $eps=1e-8$\\
        Peak learning rate    &  5e-6   \\
        Learning rate schedule & cosine decay \\
        Training epochs & 1 \\
        Training steps & 4865 \\
        Warm-up steps & 500 \\
        Global batch size & 128\\
        Gradient accumulation & 4\\
        Numerical precision & bfloat16\\

        \bottomrule
    \end{tabular}
    \vspace{0.5em}
    \label{tab:training_hype}
       \vspace{-2em}
\end{table}

\subsection{Data Generation Details}
\textbf{CoT Generation.}
In this section, we will introduce more details of CoT data generation, including the main designs and detailed prompts.

Specifically, our CoT data generation is primarily based on our annotated conversation data. Based on conversation data, we leverage diverse prompts to leverage closed-source MLLMs to generate the thinking process. A key constraint during this process is that the MLLMs are prohibited from pre-revealing the answer, ensuring that the generated CoT reflects genuine reasoning. 

Besides, the selection of MLLMs used to annotate is also different.
For visual-spatial reasoning, the input images derive from complex indoor scenes, which contain rich information. Therefore, to effectively use this information for CoT generation, we need to choose an MLLM with more creativity. Thus, for generating CoT in this task, we choose Gemini-2.0, which is more creative than GPT-4o despite being weaker in instruction following. During this process, we have some interesting observations.
For instance, when assessing the distance between a chair and a door, Gemini 2.0 adopts a practical approach. It first estimates the chair's width and then infers the distance based on how many chairs can fit between the chair and the door. In this way, these high-quality CoT data can enable VeBrain with strong spatial reasoning ability. On the other hand, GPT-4o tends to produce shorter content and relies on non-existent elements in the scene, such as floor tiles, as a basis for distance judgment. 

In the embodied CoT generation, the complexity is significantly lower compared to 3D indoor scene video data, mainly due to the limited field of view of the robot. Therefore, we adopt a more conservative strategy and used GPT-4o for CoT generation. 
Key designs here include requiring the MLLMs to think from the robot's perspective for obstacle avoidance, interaction, tracking, grasping, and destination-reaching, among other tasks. Since our control mechanism is based on a key-point policy, we visualize the key points on the input image using red circles. These marked images, combined with the QA information, are then fed into GPT-4o. This enables the robot to perform tasks such as obstacle avoidance and interaction with a high degree of reliability and interpretability. We also demonstrate the detailed prompt for generating CoT data for Complex Find tasks in locomotion.

In conclusion, our CoT generation strategy is tailored to the specific characteristics of visual spatial understanding and embodied scenes. By leveraging different models and data pre-processing techniques, we aim to enhance the interpretability and performance of VeBrain in real-world applications.

\textbf{Data composition}
Our VeBrain-600\textit{k} consists of 200\textit{k} items from multimodal understanding, 312\textit{k} items from visual-spatial reasoning, and 88\textit{k} items from self-collected robotic control. 
Specifically, the 200\textit{k} multimodal data are collected from open-source datasets, such as ShareGPT4V~\cite{dataset:sharegpt4v}  and MMInstruct~\cite{dataset:mminstruct}, \emph{etc}.
The 312\textit{k} spatial reasoning data consists of 155k items released by GPT4Scene~\cite{qi2025gpt4scene}, and 157k self-collected video data generated from ScanNet~\cite{dataset:scannet} (We actually generate 31.4\textit{k} items and repeat five times). The 88\textit{k} robotic data consists of 76\textit{k} from the legged robot and 12\textit{k} from the robotic arm. Similarly we self-collect 15.2\textit{k} and 1.2\textit{k} non-redundant data items and repeat for five and ten times for the legged robot and robotic arm respectively.

\subsection{Low-level Policy Details} 

\begin{table}[t]
    \centering
     \caption{\textbf{Policies in the Policy Pool.}}
    \footnotesize
    \resizebox{\linewidth}{!}{
    \begin{tabular}{cclcl}
\toprule
\textbf{Platform}  & \textbf{Policy} & \textbf{Description}    & \textbf{Policy} & \textbf{Description}            \\
\midrule
\multirow{7}{*}{\makecell{Legged\\Robots}} & Dump  & Pour the loaded items out of the basket.                    & Turn Right & Turn right.    \\
                               & Touch      & Lower head for humans to touch.              & Turn Left        & Turn left.                    \\
                               & Shake      & Shake hands with humans.                    & Sit   & Sit down.         \\
                               & Jump       & Jump forward.    & Wallow     & Roll left and right.         \\
                               & Scrape     & Stand on two feet and scrape.               & Lie Down   & Lie down.                    \\
                               & Squat      & Lower the body height.                      & Stand Up       & Recover to standing.                \\
                               & Heart      & Stand on two feet and make a heart sign.    & Stretch    & Stretch the body.            \\
\midrule
\multirow{2}{*}{\makecell{Robot\\Arms}}    & Grasp      & Close the robotic claw and grasp the item.  & Pull       & Pull in a specific direction. \\
                               & Release    & Open the robotic claw and release the item. &            &                    \\         
\bottomrule
    \end{tabular}}
    \vspace{0.5em}
    \label{tab:policy_pool}
       \vspace{-2em}
\end{table}

Tab.\ref{tab:policy_pool} presents all the collected actions in the policy pool, including 17 distinct implementations. With comprehensive coverage of legged robots and robot arms, this systematic organization enables robust control across different platforms.

\subsubsection{Legged Robot}

\paragraph{Setup.} The setup for legged robots consists of Unitree Go2, a RealSense D435i RGB-D camera, and a Jetson AGX Orin platform. The camera is fixed at the head of Unitree Go2, providing egocentric visual perception, and calibrated to its optical center. All perception-action pipelines operate under an edge computing paradigm, utilizing the onboard computational resources, except for MLLM inference.

\paragraph{Point Tracker.} 
In locomotion tasks, the visual observation varies as the legged robot moves, highlighting the necessity of temporal alignment between original and current keypoints. To enhance real-time performance, we adopt a series of optimizations and achieve a 15Hz execution frequency for our tracker model deployed on NVIDIA Jetson Orin. Specifically, we employ Locotrack-small\cite{LocoTrack}, a lightweight yet efficient architecture enabling near-dense point tracking. The tracking model processes two consecutive observation frames while leveraging keypoint predictions from the preceding frame for temporal consistency. 

\paragraph{Low-level Controller.} 
As described in Sec.3.2.2, the 3D coordinates for MLLM-derived keypoints in the camera frame are obtained through a series of geometric transformations. These 3D coordinates are then projected into velocity control commands $(v_x, v_y, v_{yaw})$, maintaining a mathematically constrained relationship $\displaystyle{\tan(v_{yaw}) = \frac{v_y}{v_x}}$. The velocity control commands are then executed by a low-level walking policy that generates dynamically stable motion trajectories.
\begin{table}[t]
\centering
\label{table: task define}
\scriptsize
\caption{\textbf{Definition and skill involvement of embodied tasks.}}
  \renewcommand{\arraystretch}{1.08}
  \resizebox{\linewidth}{!}{
\begin{tabular}{lllc}
\toprule
\textbf{Task Name}& &\textbf{Task Definition}  & \textbf{Skills Involved} \\
\midrule
    \rowcolor{gray!15} \multicolumn{4}{l} {$\blacktriangledown$ \emph{Locomotion Easy Tasks}} \\
\textbf{Find}    &   & Approach the target object. & Walk \\
\textbf{Track }         &       &   Approach the target object and move along with its position.   & Walk     \\
\textbf{Interaction}  &         &    Recognize human gestures, approach humans, and respond.   & Walk, Shake, Sit, Touch \\
    \rowcolor{gray!15} \multicolumn{4}{l}{$\blacktriangledown$ \emph{Locomotion Middle Tasks}} \\
\textbf{Complex Find}     &     &   Approach the target object with obstacle avoidance.   & Walk, Turn Left, Turn Right \\
\textbf{Complex Interaction }  &  &  Interaction with obstacle avoidance.  & Walk, Turn Left, Turn Right, Shake, Sit, Touch \\
\textbf{Transport}          &   & Approach to the box, and dump the object into the box. & Walk, Dump             \\
    \rowcolor{gray!15} \multicolumn{4}{l}{$\blacktriangledown$ \emph{Locomotion Hard Tasks}}   \\
\textbf{Complex Transport}     &  &  Transport with obstacle avoidance & Walk, Turn Left, Turn Right, Dump    \\
\midrule
    \rowcolor{gray!15} \multicolumn{4}{l}{$\blacktriangledown$ \emph{Manipulation Tasks}}      \\
\textbf{Banana }          &      &  Move the banana on the table to the box & Grasp, Release     \\
\textbf{Pepper}           &      &   Move the pepper on the table to the box & Grasp, Release      \\
\textbf{Carrot}           &      & Move the pepper outside of the box & Grasp, Release \\
\textbf{Kiwifruit}        &      &       Move the kiwifruit outside of the box & Grasp, Release    \\
\textbf{Open Drawer}           &      &   Pull open the half-open drawer & Pull     \\
    \rowcolor{gray!15} \multicolumn{4}{l}{$\blacktriangledown$ \emph{Manipulation Long-Horizon Tasks}}   \\
\textbf{Carrot }       &      & Open the drawer first, and move the carrot outside from the drawer. & Grasp, Release, Pull \\
\textbf{Pepper }        &      &   Open the drawer first, and move the pepper outside from the drawer. & Grasp, Release, Pull \\
\bottomrule
\end{tabular}}
\label{tab:task_define}
\end{table}

\begin{table*}[t]
    \centering
    \caption{\textbf{Ablation studies of VeBrain regarding dataset proportion, learning rate and training parameters. The default configuration on the last row best balance the three types of capabilities.}}
    \resizebox{\linewidth}{!}{
    \begin{tabular}{lcccccccc}
        \toprule
        \multirow{2}{*}{} & 
        \multicolumn{2}{c}{Multimodal Understanding} &
        \multicolumn{2}{c}{Visual-spatial Reasoning} & \multicolumn{2}{c}{Robot Control} &
        \multirow{2}{*}{Avg}\\
        
        \cmidrule(lr){2-3} \cmidrule(lr){4-5} \cmidrule(lr){6-7}
        & MMVet & MMBench & ScanQA (CIDEr) & VSI-Bench & Complex Find & Transporting & \\
        \midrule
        Qwen2.5-VL~\cite{bai2025qwen2}  & 67.1 & 83.5 & 62.7 & 35.9 & 0.0& 10.0& 43.2 \\
        \midrule

         \rowcolor{gray!15} \multicolumn{8}{l}{$\blacktriangledown$ \emph{Dataset proportion: (Default: [Multimodal, Visual-spatial Intelligence, Robot control]=[200k,312k,88k])}}   \\
         400k$,$312k$,$88k & 72.8 & 84.3 & 97.8   & 38.5 & 70.0 & 85.0 & 74.7\\
         200k$,$400k$,$88k & 71.5 & 83.5 & 101.6 & 40.2 & 75.0 & 85.0 & 76.1\\
         200k$,$312k$,$200k & 69.9 & 83.1 & 98.4 & 39.1 & 85.0 & 90.0 & 77.6\\
        \rowcolor{gray!15} \multicolumn{8}{l}{$\blacktriangledown$ \emph{Learning rate: (Default: 5e-6)}}   \\
         2e-6 & 70.4 & 83.3 & 99.8 & 39.2 & 75.0 & 80.0 & 74.6\\
         1e-5 & 43.5 & 70.2 & 90.2 & 36.9 & 80.0 & 90.0 & 68.5\\
         \rowcolor{gray!15} \multicolumn{8}{l}{$\blacktriangledown$ \emph{Training Manner: (Default: Fully finetuning)}}   \\
         LoRA Adapter & 71.3 & 83.5 & 97.6 & 38.7 & 80.0 & 90.0 & 76.9\\

        \midrule
        VeBrain (Default) & 72.7 & 83.7 & 101.5 & 39.9 & 80.0 & 90.0 & 78.0 \\
        \bottomrule
    \end{tabular}}
    \label{table: appendix_abla} 
    \vspace{-1.em}
\end{table*}

\subsubsection{Robot Arm}

\paragraph{Setup.} The robot setup consists of a 7-DoF tabletop Franka Emika Panda robot arm equipped with a Robotiq 2F-85 gripper. A RealSense D435i RGB-D camera, positioned approximately 1.5m away from the robot in a third-person view, captures RGB-D scenes at each inference timestamp. The camera is calibrated with the origin of the world coordinate frame is aligned with the base of the robot arm. Robot control is managed from a desktop computer running ROS.

\paragraph{Robot Control.} Upon acquiring the RGB image $\mathcal{I}_c$ and the depth image $\mathcal{I}_d$ captured by the camera, $\mathcal{I}_c$ together with prompt are fed into VeBrain to predict 2 target points, $p_1$ and $p_2$, which correspond to the intended antipodal grasping positions for the gripper at each inference. To determine the target pose of the robot arm, the translation component is obtained by querying the depth value at the midpoint $p*$ of $p_1$ and $p_2$ in $\mathcal{I}_d$, and subsequently transforming this pixel location into the world coordinate frame. As for rotation, we adopt a top-down grasping paradigm, which requires only adjusting the gripper's orientation around its $z$-axis according to the direction defined by $p_1$ and $p_2$. The computed 6D end-effector pose is then passed to the MoveIt library\footnote{https://github.com/moveit/moveit} for motion planning. The gripper's closing width is determined by a predefined force threshold. Notably, when the robot task involves opening a drawer, the gripper skips the closing action.

\subsection{Robotic Task Details}
In this paper, we design hierarchical embodied tasks for both locomotion and manipulation. The definitions of each task are shown in Tab.~\ref{tab:task_define}. For locomotion, the Easy Tasks include three key tasks: find, track, and interaction. These tasks evaluate the basic abilities of VeBrain-equipped legged robots in static object searching, dynamic following, and human intention recognition, respectively. The Interaction task involves four gestures: come, sit, shake, and touch. In the Middle Tasks, we introduce obstacle avoidance into the find and interaction tasks, and add a new task, transport, to assess VeBrain's practical application capabilities. Finally, the Hard Locomotion Tasks involve complex transport with obstacle avoidance, further challenging the robot's mobility and coordination in dynamic environments.

For manipulation, we also design two levels of tasks: normal tasks and long-horizon tasks. The normal manipulation tasks include moving objects like bananas, peppers, and kiwis to specific locations, \textit{e.g.,} from the table to the box or outside the box, as well as pulling open a half-open drawer. These tasks test the basic manipulation skills, such as grasping and releasing. The long-horizon manipulation tasks require performing multiple-step sequences, such as opening the drawer first and then moving the carrot or pepper outside of it. These tasks assess the robot's ability to execute complex, multi-step actions and demonstrate precise control over its manipulation process.

In our evaluation, all tasks, except for the Interaction and Complex Interaction tasks, will be tested 10 times to ensure reliability. For the Interaction task, each of the four gestures (come, sit, shake, touch) will be tested 5 times to assess the robot's responsiveness and accuracy in recognizing human gestures. Moreover, the object layout in the scene will be slightly modified in each test.

\section{More Results}
\subsection{Ablation Study}

In Tab. \ref{table: appendix_abla}, we further conduct ablation studies from three perspectives. 1) For the dataset proportion, our default VeBrain-600\textit{k} comprises 200\textit{k} from multimodal conversations, 312\textit{k} from visual-spatial reasoning, and 88\textit{k} from self-collected robotic dataset. We respectively increase the amount of data items for the three types of tasks, and found that the corresponding capability increases slightly but at the cost of the other two capabilities. Therefore, we maintain the original data proportion for the best trade-off among the three types of capabilities. 
2) For the learning rate, we adopt the peak learning rate of 5e-6. A lower learning rate, like 2e-6, results in marginal degradation across all three capabilities, while a higher learning rate, such as 1e-5, leads to a remarkable decline in multimodal understanding performance.
3) If we optimize VeBrain via LoRA adapter rather than fully optimize the LLM, the overall performance also shrinks slightly due to the less learnable parameters.
The ablations for the above three perspectives further substantiate the superiority of training configurations in our default setting.

\end{document}